\documentclass[letterpaper]{article} % DO NOT CHANGE THIS
\usepackage{aaai2026}  % DO NOT CHANGE THIS
\usepackage{times}  % DO NOT CHANGE THIS
\usepackage{helvet}  % DO NOT CHANGE THIS
\usepackage{courier}  % DO NOT CHANGE THIS
\usepackage[hyphens]{url}  % DO NOT CHANGE THIS
\usepackage{graphicx} % DO NOT CHANGE THIS
\usepackage{natbib}  % DO NOT CHANGE THIS AND DO NOT ADD ANY OPTIONS TO IT
\usepackage{caption} % DO NOT CHANGE THIS AND DO NOT ADD ANY OPTIONS TO IT
\usepackage{algorithm}
\usepackage{booktabs}
\usepackage{newfloat}
\usepackage{listings}
\DeclareCaptionStyle{ruled}{labelfont=normalfont,labelsep=colon,strut=off} % DO NOT CHANGE THIS
\floatstyle{ruled}
\newfloat{listing}{tb}{lst}{}
\floatname{listing}{Listing}
\usepackage{graphicx}

\usepackage[accsupp]{axessibility}  % Improves PDF readability for those with disabilities.

\usepackage[utf8]{inputenc} % allow utf-8 input
\usepackage[T1]{fontenc}    % use 8-bit T1 fonts

\usepackage{url}            % simple URL typesetting
\usepackage{booktabs}       % professional-quality tables
\usepackage{amsfonts}       % blackboard math symbols
\usepackage{nicefrac}       % compact symbols for 1/2, etc.
\usepackage{microtype}      % microtypography
\usepackage{xcolor}         % colors

\usepackage{algpseudocode}

\usepackage{xspace}
\usepackage{float}
\usepackage{amsmath, amssymb, bm, dsfont}
\usepackage{mathtools}
\usepackage{multirow}
\usepackage{comment}
\usepackage{lipsum,booktabs}
\usepackage{caption}
\usepackage{subcaption}
\newcommand{\matr}[1]{\mathbf{#1}}
\newcommand{\vect}[1]{\mathbf{#1}} 

\usepackage{cleveref} % Added for \cref support
\newtheorem{theorem}{Theorem}

\usepackage{calc}     % For \widthof

\newcommand{\system}{STRIDE\xspace}
\title{\system: Structure and Embedding Distillation with Attention for Graph Neural Networks}

\author{
  \normalsize Anshul Ahluwalia\equalcontrib,
    Payman Behnam\equalcontrib,
    Rohit Das,
    Alind Khare,
    Biswadeep Chakraborty,
    Pan Li, 
    Alexey Tumanov
}
\affiliations{
    Georgia Institute of Technology\\
    \{aahluwalia30, payman.behnam, rohdas, akhare39, biswadeep, panli,
atumanov\}@gatech.edu
}

\begin{document}
\maketitle

\begin{abstract}
Recent advancements in Graph Neural Networks (GNNs) have led to increased model sizes to enhance their capacity and accuracy.
Such large models incur high memory usage, latency, and computational costs, thereby restricting their inference deployment.  
GNN compression techniques compress large  GNNs into smaller ones with negligible accuracy loss. One of the most promising compression techniques is knowledge distillation (KD). However, most KD approaches for GNNs only consider the outputs of the last layers and do not consider the outputs of the intermediate layers of the GNNs. The intermediate layers may contain important inductive biases indicated by the graph structure and embeddings. Ignoring these layers may lead to a high accuracy drop, especially when the compression ratio is high. To address these shortcomings, we propose a novel KD approach for GNN compression that we call Structure and Embedding Distillation with Attention (\system). \system utilizes attention to identify important intermediate teacher-student layer pairs and focuses on using those pairs to align graph structure and node embeddings.
We evaluate \system on several datasets such as OGBN-Mag and OGBN-Arxiv using different
model architectures including GCNIIs, RGCNs, and GraphSAGE. On average, \system achieves a $2.13\%$ increase in accuracy with a $32.3\times$ compression ratio on OGBN-Mag, a large graph dataset, compared to state-of-the-art approaches. On smaller datasets (e.g., Pubmed), \system achieves up to a ${141\times}$ compression ratio with the same accuracy as state-of-the-art approaches. These results highlight the effectiveness of focusing on intermediate-layer knowledge to obtain compact, accurate, and practical GNN models.

\end{abstract}

\section{Introduction}
\label{sec:intro}
The rapid growth in the scale and complexity of real-world graphs, including social networks~\cite{MAG_2020}, web graphs~\cite{hyperlinkgraph}, knowledge graphs~\cite{dbpedia}, e-commerce graphs~\cite{graphexplorer}, and biological networks~\cite{esfahani2023ms} has driven significant advancements in Graph Neural Network (GNN) architectures, making them increasingly deeper and more expressive.

Recent studies examining neural scaling laws demonstrate notable accuracy improvements for GNNs by increasing depth, parameter count, and training dataset size~\cite{li2021graph, yang2022mgraphdta, liu2024neural, sypetkowski2024scalability, ma2022graph, chen2024hopgnn, airas2025scaling}. For instance, Liu et al.~\cite{liu2024neural} show clear performance gains with deeper and wider GNN models. Similarly, Sypetkowski et al.~\cite{sypetkowski2024scalability} highlight enhanced performance in molecular graph tasks by employing larger models and richer pretraining datasets.

However, these performance gains come at significant costs, including increased computational complexity, memory usage, storage requirements, over-smoothing, and over-squashing, which complicate their practical deployment~\cite{liu2024scalable, sypetkowski2024scalability, di2023over}. The growing demand for real-time inference further exacerbates these deployment challenges. Real-time applications such as autonomous vehicle point cloud segmentation~\cite{shi2020point,sarkar2023flowgnn}, high-energy physics data acquisition~\cite{shlomi2020graph}, real-time recommendation systems~\cite{liu_2022},rapid image retrieval~\cite{Formal_2020}, and spam detection~\cite{Li_2019} require extremely low inference latencies. Unfortunately, as GNN model complexity increases, inference latency escalates sharply, leading to substantial practical deployment barriers~\cite{zhou2021accelerating, que_2023, tan_2023, huang2021scaling, kiningham2022grip}. 
Consequently, compressing large GNNs into smaller, low-latency models without losing accuracy is now a key research goal.

Knowledge Distillation (KD) is a widely adopted model compression technique in which a compact \textit{student} model is trained using supervision signals from a larger, well-performing \textit{teacher} model~\cite {hinton2015distilling}. Although conventional KD can be applied directly to GNNs, it largely ignores structural properties inherent in graphs (e.g., Fitnets~\cite{fitnetsromero2015} and Attention Transfer (AT)~\cite{AT_2016}). Hence, simply matching node embeddings or attention maps overlooks critical structural information in GNNs, limiting the effectiveness of these methods when directly applied to graph data. 

Recently, Tian et al.~\cite{tian2023knowledge} identify three primary types of transferable knowledge in GNN distillation: \textit{logits}, \textit{structure}, and \textit{embeddings}. Among these, logits-based distillation using soft-label predictions is straightforward, prompting recent research to focus on more advanced methods for transferring structural and embedding knowledge from teacher to student GNNs. 
Structural knowledge captures how nodes are interconnected and how the teacher network encodes graph topology~\cite{lspyang2021distilling}. Embedding knowledge mainly reflects node-level semantic relationships in the learned feature space~\cite{he2022compressing, globalkdJoshi_2022}.
Early KD methods for GNNs primarily focused on preserving local graph structure. For example, LSP~\cite{lspyang2021distilling} emphasizes the local structural alignment between the teacher and the student. 
Joshi et al. build on LSP by introducing GSP, which distills knowledge using all pairwise node similarities, and G-CRD, which preserves global topology via contrastive alignment of student and teacher node features \cite{globalkdJoshi_2022}. Later, GraphAKD~\cite{he2022compressing} directly distills embedding knowledge by forcing the student’s node and class-level embeddings to match those of the teacher through adversarial training.

On the other hand, several studies have developed attention mechanisms for KD in GNNs, typically focusing on transferring knowledge from multiple teachers to a single student~\cite{MulDE_2021, Zhang_2022} or leveraging only embedding or structural features to enhance distillation. 

Despite this progress, current KD methods for GNNs remain fundamentally limited by focusing mainly on final-layer embeddings, neglecting valuable information captured in intermediate layers~\cite{Baxter_2000, uselisintermediate-2025}. Intermediate GNN layers encode distinct graph connectivity patterns and hierarchical structural relationships, which are critical for generalization. Ignoring these intermediate representations restricts the student's capability to learn deeper structural relationships, causing it to rely heavily on superficial mappings between node attributes and final-layer outputs, thus hindering generalization to unseen graph data. In particular, \textbf{it is essential to jointly leverage structural relationships, node embeddings, and intermediate-layer representations. These components collectively encode distinct yet complementary information about graph data.}

Unfortunately, aligning intermediate-layer representations poses a nontrivial challenge due to inherent architectural differences between teacher and student models. Typically, compressed student networks contain fewer layers, creating a mismatch in intermediate representations and preventing straightforward one-to-one alignment. Consequently, most existing GNN distillation methods avoid intermediate-layer alignment, limiting their ability to fully utilize the rich hierarchical information embedded within teacher layers ~\cite{ globalkdJoshi_2022, kim2021compressing, amalgamateYongcheng2021cvpr, wang2021online, huo2023t2, wang2024lightweight}. Addressing this challenge represents an important research frontier in GNN distillation, motivating innovative approaches to dynamically align intermediate representations without relying on fixed-layer correspondences.

To address the shortcomings of existing KD methods for Graph Neural Networks, we propose \textit{Structure and Embedding Distillation with Attention} (\system). The core novelty of our approach lies in using a trainable attention mechanism to automatically identify and align the most informative pairs of intermediate layers, resolving the longstanding challenge of mismatched architectures in GNN distillation. Unlike prior methods that rely on explicit, fixed layer mappings, \system enables flexible and dynamic distillation of structural and embedding information even when the teacher and student networks differ significantly in depth and architecture.

\begin{figure}
    \centering \includegraphics[width=0.5\textwidth]{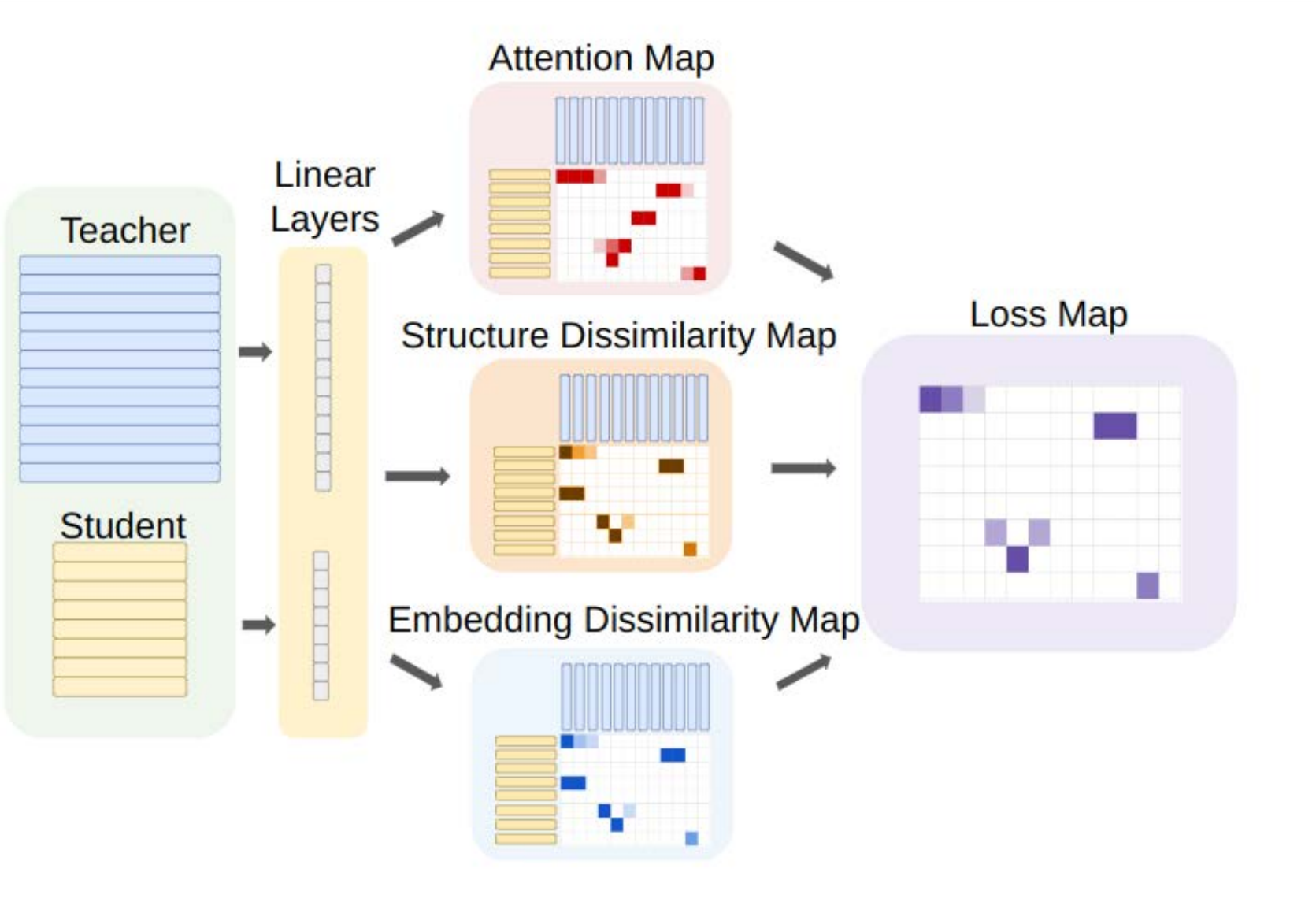}
    \vspace{-6ex}
    \caption{
        \system generates an attention map using a trainable attention mechanism and a dissimilarity map using a trainable subspace projection. The loss matrix is an element-wise multiplication of the attention matrix and the dissimilarity matrix. 
        \vspace{-2ex}
    \label{fig:abkd}}
    \vspace{-10pt}
\end{figure}

Specifically, \system projects intermediate hidden representations from both teacher and student GNNs into a shared latent space, facilitating meaningful comparison across layers. Subsequently, a learned attention mechanism dynamically weighs the importance of aligning each potential pair of teacher and student layers based on their representational similarity and informativeness. By aligning embeddings and structural information at multiple intermediate layers, \system encourages the student network to internalize the hierarchical reasoning and richer graph structures captured by the teacher, rather than relying solely on superficial input-output mappings (see Figure~\ref{fig:abkd}).

The main contributions of our work include:
\begin{enumerate}
    \item We introduce \system, the first attention-based GNN knowledge distillation framework capable of \textbf{simultaneously aligning  structural and embedding} representations across \textbf{all intermediate layers}. Crucially, \system accommodates substantial architectural differences (e.g., depth, hidden dimensions) between teacher and student models, without requiring explicit layer correspondence.
    
    \item We develop a novel \textbf{attention-driven alignment mechanism}, enabling dynamic identification of critical teacher-student layer pairs. This facilitates effective knowledge transfer and improves the student’s representational capability across diverse GNN configurations.
    
    \item We provide extensive empirical validation of \system using multiple widely-adopted benchmark datasets, such as OGBN-Mag~\cite{MAG_2020} and OGBN-Arxiv~\cite{MAG_2020}, and various GNN architectures including GCNII~\cite{Chen_2020}, RGCN~\cite{RGCN_2017}, and GraphSAGE~\cite{GraphSAGE_2017}. Our experiments demonstrate consistent performance improvements in accuracy and generalization across different degrees of model compression. On the large-scale OGBN-Mag dataset, our method outperforms state-of-the-art approaches by 2.13\% in accuracy and achieves a $32.3\times$ compression ratio.
    On smaller datasets (e.g., Pubmed), \system achieves up to a $141\times$ compression ratio with the same accuracy as state-of-the-art approaches.
\end{enumerate}

\section{Proposed Approach}
\label{sec:approach}

\label{sec:approach:prelim}

\subsection{Intuition and Mathematical Foundations}
In this section, we first discuss the intuition behind \system and introduce some of the mathematical definitions needed to explain it thoroughly.
The mathematical notations are explained in the Appendix of the paper.
\subsubsection{SoftKD Intuition}
\label{sec:approach:prelim:softkd}

In SoftKD \cite{hinton2015distilling}, we compute two different losses. The first, $H(s_p, y)$, is a cross-entropy loss between the output student probability distribution and the ground truth labels. The other, $H(s_p, t_p)$, is a cross-entropy loss between the output student probability distribution and the output teacher probability distribution. The total loss is defined as:  

\vspace{-2ex}
\begin{equation}
    L_{KD} = H(s_p, y) + \alpha H(s_p, t_p)
\end{equation} 

Here, $\alpha$ is a hyper-parameter controlling how much the KD loss affects the total loss. The goal is to align the output student probability distribution with the output teacher probability distribution. The higher $H(s_p, t_p)$ is, the less aligned the student and teacher output probability distributions are. 

\subsubsection{\system Intuition}
\label{sec:approach:prelim:abkd}
Similarly, \system aims to incorporate alignment, but goes beyond final output alignment by focusing on intermediate layers, which encode valuable inductive biases. We pay special attention to the structural information within these representations. A key challenge arises from our goal to support arbitrary teacher-student architectures. Since their number of layers may differ, a direct layer-to-layer alignment is not feasible.

\system solves this problem by identifying which teacher-student layer pairs are the most important to align via an attention mechanism. This mechanism works with an arbitrary number of teacher and student layers, which makes this approach amenable to any arbitrary teacher-student configuration. \system also proposes a reprojection technique to account for the student and teacher networks having different hidden dimensions. The output of each hidden layer for both the teacher and student networks is projected into a standardized embedding dimension, which ensures that it will work with student and teacher networks of any embedding dimension (Figure~\ref{fig:abkd}). 

As each layer represents unique semantic information, an important challenge is to ensure that each layer's feature map is not smoothed out by a single projection matrix. To this end, we use separate trainable linear layers for each hidden layer in both the teacher and student networks to ensure that we do not lose any valuable semantic information in the hidden layers. These trainable linear layers help us construct the three key components of \system, which are the attention map, the structural dissimilarity map, and the embedding dissimilarity map. At a high level, the attention map tells us how important each teacher-student layer pair is, while the dissimilarity maps tell us how distant the feature maps of each teacher-student layer pair are in terms of both embedding and structure. The teacher-student layer pairs with higher attention scores are deemed as more important; \system focuses on reducing their structural and embedding dissimilarity scores during training (Figure~\ref{fig:abkd}). 

\subsubsection{Mathematical Foundation}
\label{sec:approach:prelim:math}
Without loss of generality, we consider distilling a general Graph Convolutional Network (GCN) \cite{KipfW16}, in which the output of the $l$-th layer is:
\vspace{-2ex}
\begin{equation}
    {\matr{H}^{(l)} = \sigma(\hat{\matr{A}}\matr{H}^{(l-1)}\matr{W}^{(l)})}
\end{equation}
Here, $\sigma$ is an activation function. The $\hat{\matr{A}} = \matr{D}^{-1/2}\matr{A}\matr{D}^{-1/2}$ is the normalized adjacency matrix, where $\matr{A}$ is the adjacency matrix and $\matr{D}$ is the diagonal degree matrix. The term $\matr{H}^{(l-1)} \in \mathbb{R}^{n \times d_{l-1}}$ represents the node feature matrix from the previous layer, and $\matr{W}^{(l)} \in \mathbb{R}^{d_{l-1} \times d_l}$ represents the trainable weight matrix of the current layer. Note that this formulation allows the hidden dimension to change among layers.

For STRIDE, we are interested in the collection of intermediate feature representations from both the teacher and student networks. Let us define the teacher network $\mathcal{T}$ and the student network $\mathcal{S}$ as having $T_l$ and $S_l$ layers, respectively. We collect the \textit{pre-activation} feature maps from each layer.

Let $\matr{T}_i \in \mathbb{R}^{n \times d_i^t}$ be the pre-activation output of the $i$-th layer of the teacher network, where $n$ is the number of nodes and $d_i^t$ is the feature dimension of that specific layer. Similarly, let $\matr{S}_j \in \mathbb{R}^{n \times d_j^s}$ be the pre-activation output of the $j$-th layer of the student network with output dimension $d_j^s$. Our goal is to distill knowledge from the set of teacher representations $\{\matr{T}_i\}_{i=1}^{T_l}$ to the set of student representations $\{\matr{S}_j\}_{j=1}^{S_l}$.

\subsection{\system Mechanism}
\label{sec:approach:abkd}

\subsubsection{Attention Scores}
\label{sec:approach:abkd:attention}

The first step of \system is to generate the attention matrix $\matr{\alpha} \in \mathbb{R}^{T_l \times S_l}$. An element $\alpha_{ij}$  represents an ``importance'' score for the layer pair consisting of teacher layer $i$ and student layer $j$. We take the average of the feature maps along the node dimension to compute a mean node feature for every layer in both the teacher and student networks. We call these tensors $\matr{T}_a \in \mathbb{R}^{T_l \times d_t}$ and $\matr{S}_a \in \mathbb{R}^{S_l \times d_s}$. Then, we pass each layer in $\matr{T}_a$ through its own linear layer to create $\matr{T}_p \in \mathbb{R}^{T_l \times d_a}$, where $d_a$ is the embedding dimension of \system. Similarly, we create $\matr{S}_p \in \mathbb{R}^{S_l \times d_a}$. We can finally generate $\matr{\alpha}$ in the following manner:
\vspace{-1ex}
\begin{equation}
    \matr{\alpha} = \text{softmax}\left(\frac{\matr{T}_p\matr{S}_p^T}{\sqrt{d_a}}\right)
\end{equation}

\subsubsection{Embedding Dissimilarity Scores}
\label{sec:approach:abkd:embedding:dissimilarity}
The next step is to compute a pairwise embedding dissimilarity score for each teacher-student layer pair. Again, we project the features into $d_a$. For calculating the attention scores, we average over the node dimension before projecting, as our goal was to identify important layers. When calculating the pairwise embedding dissimilarity, we want to incorporate the per-node embeddings. So, we use a separate set of projection matrices. We use $\matr{P}_t \in \mathbb{R}^{d_t \times d_a}$ and $\matr{P}_s \in \mathbb{R}^{d_s \times d_a}$ to represent the projections.
However, distance metrics are less semantically valuable if $d_a$ is high. To alleviate this problem, we define a trainable matrix $\matr{P} \in \mathbb{R}^{d_a \times d_a}$ to project all vectors into the subspace defined by the column space of $\matr{P}$. Since the rank of $\matr{P}$ can be less than $d_a$, distance metrics within the learned subspace can be more semantically valuable.

The final step is to average over the embedding dimension and then produce the embedding dissimilarity matrix $\matr{D}_{\text{emb}} \in \mathbb{R}^{T_l \times S_l}$. Its elements give the dissimilarity scores for each teacher-student layer pair. To calculate the embedding dissimilarity, we experiment with Euclidean and cosine distance, but Euclidean distance generally tends to perform better. The embedding dissimilarity score for a layer pair $(i, j)$ can be represented as:
\vspace{-2ex}
\begin{equation}
    (\matr{D}_{\text{emb}})_{ij} = \left\|(\matr{T}_i\matr{P}_t - \matr{S}_j\matr{P}_s)\matr{P}\frac{\vect{1}_{d_a}}{d_a}\right\|_2^2
\end{equation}
where $\matr{T}_i$ and $\matr{S}_j$ are the node feature maps for the respective layers, and $\vect{1}_{d_a}$ is a vector of ones in $\mathbb{R}^{d_a}$.

\subsubsection{Structural Dissimilarity Scores}\label{sec:approach:abkd:structural:dissimilarity}
Before calculating the loss, the final step is to compute a pairwise structural dissimilarity for each teacher-student layer pair. As usual, we first project teacher and student features to $d_a$ via a trainable linear projection. We use $\matr{E}_t \in \mathbb{R}^{d_t \times d_a}$ and $\matr{E}_s \in \mathbb{R}^{d_s \times d_a}$ to represent these trainable projections. After projecting to $d_a$ to find the structural dissimilarity difference, we simply use the G-CRD loss~\cite{globalkdJoshi_2022}. We can also use other structure-aligning losses, such as LSP or GSP, but experimentally we find that using G-CRD produces the best results. We use the G-CRD loss to find the structural dissimilarity for every teacher-student layer pair to produce the structural dissimilarity matrix $\matr{D}_{\text{str}} \in \mathbb{R}^{T_l \times S_l}$. The structural dissimilarity score for a layer pair $(i, j)$ can be represented as:
\begin{equation}
    (\matr{D}_{\text{str}})_{ij} = \phi(\matr{T}_i\matr{E}_t, \matr{S}_j\matr{E}_s)
\end{equation}
where $\phi$ represents the structure-aligning loss of G-CRD.

\subsubsection{Final Loss Calculation}
\label{sec:approach:abkd:loss}

To produce the final loss value, we first compute a total dissimilarity matrix $\matr{M} = \matr{D}_{\text{str}} + \matr{D}_{\text{emb}}$. We then element-wise multiply it with the attention matrix $\matr{\alpha}$ and take the mean to produce a single number that represents the \system distillation loss, $L_{STRIDE}$.
\vspace{-1ex}
\begin{equation}
    L_{STRIDE} = (\vect{1}_{T_l})^T(\matr{\alpha} \odot \matr{M})\left(\frac{\vect{1}_{S_l}}{S_l}\right)
    \label{eq:abkd}
\end{equation}
The final loss is calculated as:
\begin{equation}
    L = \mathcal{H}(s_p, y) + \beta L_{STRIDE}
\end{equation}
where $\mathcal{H}(s_p, y)$ is the standard cross-entropy loss between the student's predictions and the ground truth labels. There is one important theorem to consider that proves $L_{STRIDE}$ distills valuable knowledge from the teacher network to the student network.

\begin{theorem}[\system Cross-Layer Gradient Dependence]
\label{thm:cross_layer_grad}
Let the STRIDE distillation loss be $L_{STRIDE}$, which is a function of the set of all teacher weight matrices $\{\matr{W}_i^t\}_{i=1}^{T_l}$ and student weight matrices $\{\matr{W}_j^s\}_{j=1}^{S_l}$. The gradient of the loss with respect to any student layer's weight matrix, $\matr{W}_j^s$, is functionally dependent on every teacher layer's weight matrix, $\matr{W}_i^t$. Formally:

\begin{align}
\frac{\partial L_{STRIDE}}{\partial \matr{W}_j^s} &= f(\{\matr{W}_k^t\}_{k=1}^{T_l}, \{\matr{W}_l^s\}_{l=1}^{S_l}) \nonumber \\
&\quad \forall i \in [1, T_l], \forall j \in [1, S_l]
\end{align}

This holds even for teacher layers $i$ that are deeper than the student layer $j$ (i.e., $i > j$).
\end{theorem}

\paragraph{Intuitive Proof} The full proof involves a detailed expansion of the partial derivatives and is provided in the Appendix. The core intuition, however, is straightforward and relies on the chain rule through the attention mechanism.
\begin{enumerate}
    \item The total STRIDE loss is a sum of losses for each teacher-student layer pair $(i,j)$, weighted by an attention score $\alpha_{ij}$. The loss for a single pair is $L_{ij} = \alpha_{ij} \cdot M_{ij}$, where $M_{ij}$ is the dissimilarity score.

    \item Crucially, the \textbf{attention score $\alpha_{ij}$ is a function of the outputs} of teacher layer $i$ (denoted $\matr{H}_i^t$) and student layer $j$ (denoted $\matr{H}_j^s$).
    % $$
    % \alpha_{ij} \propto g(\matr{H}_i^t, \matr{H}_j^s)
    % $$
    
    \begin{align}
    \alpha_{ij} &\propto `g(\matr{H}_i^t, \matr{H}_j^s)
    \end{align}
    \item The output of any teacher layer, $\matr{H}_i^t$, is a function of its weights, $\matr{H}_i^t = f_t(\matr{W}_1^t, \dots, \matr{W}_i^t)$. Likewise, the student's output $\matr{H}_j^s$ is a function of its weights, $\matr{H}_j^s = f_s(\matr{W}_1^s, \dots, \matr{W}_j^s)$.

    \item Therefore, when calculating the weight update for $\matr{W}_j^s$ via the gradient $\frac{\partial L_{STRIDE}}{\partial \matr{W}_j^s}$, the chain rule must backpropagate through $\alpha_{ij}$. Since $\alpha_{ij}$ directly depends on the teacher's output $\matr{H}_i^t$, the gradient flowing to the student weight $\matr{W}_j^s$ will necessarily contain terms involving the teacher's weight $\matr{W}_i^t$.
\end{enumerate}
This structure creates a computational graph where the teacher's weights from \textit{every} layer influence the gradient of \textit{every} student layer, thus proving the cross-layer dependency.

\paragraph{Main Takeaway} Theorem~\ref{thm:cross_layer_grad} provides the theoretical justification for our core claim: STRIDE enables a richer, more comprehensive knowledge transfer than prior methods. The key insight is that our attention mechanism creates \textbf{direct gradient pathways from all teacher layers to all student layers}. This means a shallow student layer (e.g., layer 1) can receive immediate supervisory signals not just from the teacher's first layer, but also from its deepest layers (e.g., layer 5). This allows the student to learn how to represent complex, higher-order neighborhood information---a task usually reserved for deeper layers---much earlier in its own architecture. This ability to distill the teacher's entire representational hierarchy into a more compact student model is what leads to the significant gains in accuracy and generalization that we observe in our experiments.

\section{Experiments}
\label{sec:experiments}
\subsection{Experimental Setup}
\label{sec:experiments:setup}
For our main experiments, we test \system on two difficult datasets: OGBN-Mag and OGBN-Arxiv \cite{OGBN_2020, MAG_2020}. These datasets utilize temporal splitting to create validation and test sets that assess a model's ability to generalize to out-of-distribution data. For OGBN-Mag, we run experiments using RGCN~\cite{RGCN_2017} as the teacher and student models, and for OGBN-Arxiv, we run experiments using GAT~\cite{veličković_2018} as the teacher model and GraphSAGE~\cite{GraphSAGE_2017} as the student model.
We also test \system on Cora \cite{Mccallum_2000}, Citeseer \cite{Namata_2008}, Pubmed \cite{Namata_2012} and NELL \cite{Betteridge_2010}. 

This settings allows us to evaluate the effectiveness of \system for different GNN architectures and datasets. It also allows us to assess if \system can distill information between different types of GNN architectures. In our experiments, we keep the teacher model architecture and weights fixed and only modify the size of the student network. Each distillation method starts from the same set of weights and trains for the same number of epochs across 5 runs. For our baselines, we consider LSP \cite{lspyang2021distilling}, GSP \cite{globalkdJoshi_2022}, G-CRD \cite{globalkdJoshi_2022}, Fitnets \cite{fitnetsromero2015}, and Attention Transfer (AT)~\cite{AT_2016},  which are most closely related to \system. We run all experiments on a Tesla V100 GPU.

\subsection{Experimental Results}
\label{sec:experiments:results}

\subsubsection{Out-Of-Distribution Evaluation}
\begin{figure}[h]
    \centering
    \includegraphics[width=0.60\columnwidth]{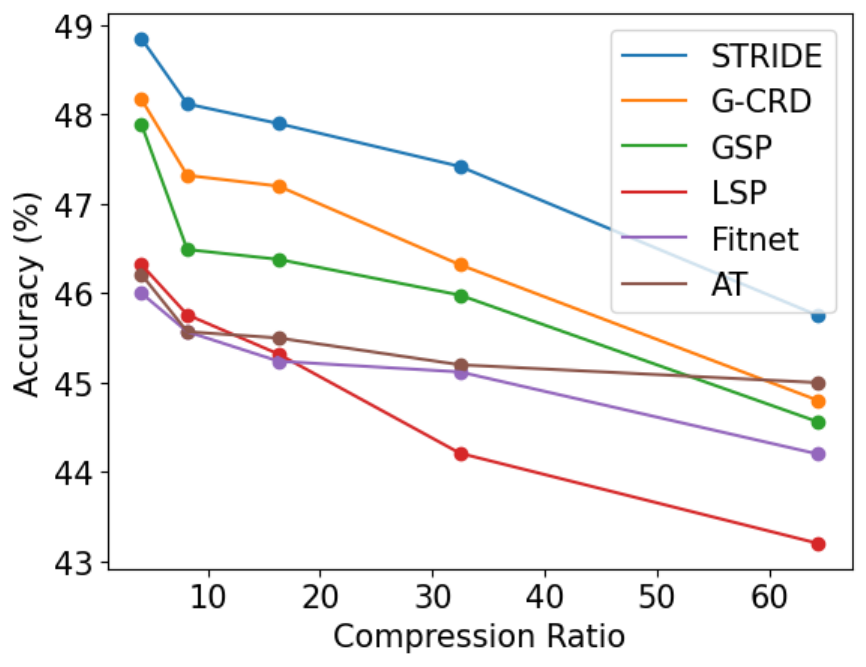}
    \vspace{-2ex}
    \caption{Test Accuracy vs. Compression Ratio:  A comparison of KD methods applied to student models of different sizes trained on the OGBN-Mag dataset. The teacher model was the same as the one described in Table~\ref{table:avgacc-main}. The student model was a two-layer RGCN, and we varied the embedding dimension from 16 to 512 to induce this Pareto frontier.}
    \label{fig:acc-compression}
    %\vspace{-1ex}
\end{figure}

As shown in Table~\ref{table:avgacc-main} and Figure~\ref{fig:acc-compression}, \system consistently outperforms SOTA across various compression ratios on OGBN-Mag and OGBN-Arxiv. Notably, it achieves gains of $2.13\%$ and $1.70\%$ at $32.3\times$ and $16.1\times$ compression, respectively. Since these benchmarks target out-of-distribution generalization, the results demonstrate \system's ability to produce student models with superior generalization compared to existing KD methods.

\vspace{-1ex}
\begin{table}[!ht]
    \centering 
        \resizebox{0.48\textwidth}{!}{
        \begin{tabular}{r|ccc}
            \hline
             {Dataset} & {OGBN-Mag} & {OGBN-Arxiv}  \\
            {Teacher} & {RGCN (3L-512H-5.5M)} & {GAT (3L-750H-1.4M)} \\
            {Student} & {RGCN (2L-32H-170K)} & {GraphSAGE (2L-256H-87K)} \\
            \hline
             Teacher 
                & $49.80$
                & $74.20$ \\
              \hline
             Student 
                & $44.23 \pm 0.47$
                & $70.87 \pm 0.58$ \\
            Fitnets 
                & $44.87 \pm 0.84$
                & $71.32 \pm 0.32$ \\
             AT 
                & $43.87 \pm 0.67$
                & $71.04 \pm 0.48$ \\ 
             LSP
                & $45.21 \pm 0.54$
                & $71.47 \pm 0.45$ \\
             GSP
                & $44.97 \pm 0.58$
                & $71.97 \pm 0.64$ \\
             G-CRD
                & $45.42 \pm 0.43$
                & $71.87 \pm 0.56$ \\
            \system
                & $\mathbf{47.55 \pm 0.28}$
                & $\mathbf{73.67 \pm 0.49}$ \\
            \hline
             Ratio 
                & $32.3 \times$ 
                & $16.1\times$ \\
            \hline
        \end{tabular}
        }
    \vspace{-2ex}
    \caption{
    Average accuracies for a variety of large datasets.
    The results are based on the average of five trials, with each distillation method applied to the same set of student weights. The notation aL-bH-cM means the model has ``a'' layers, a hidden dimension of ``b'', and ``c'' million trainable parameters.}\label{table:avgacc-main}
\end{table}

\vspace{-3ex}

\begin{figure}[h]
   \centering
   \vspace{-2ex}
\includegraphics[width=0.70\columnwidth]{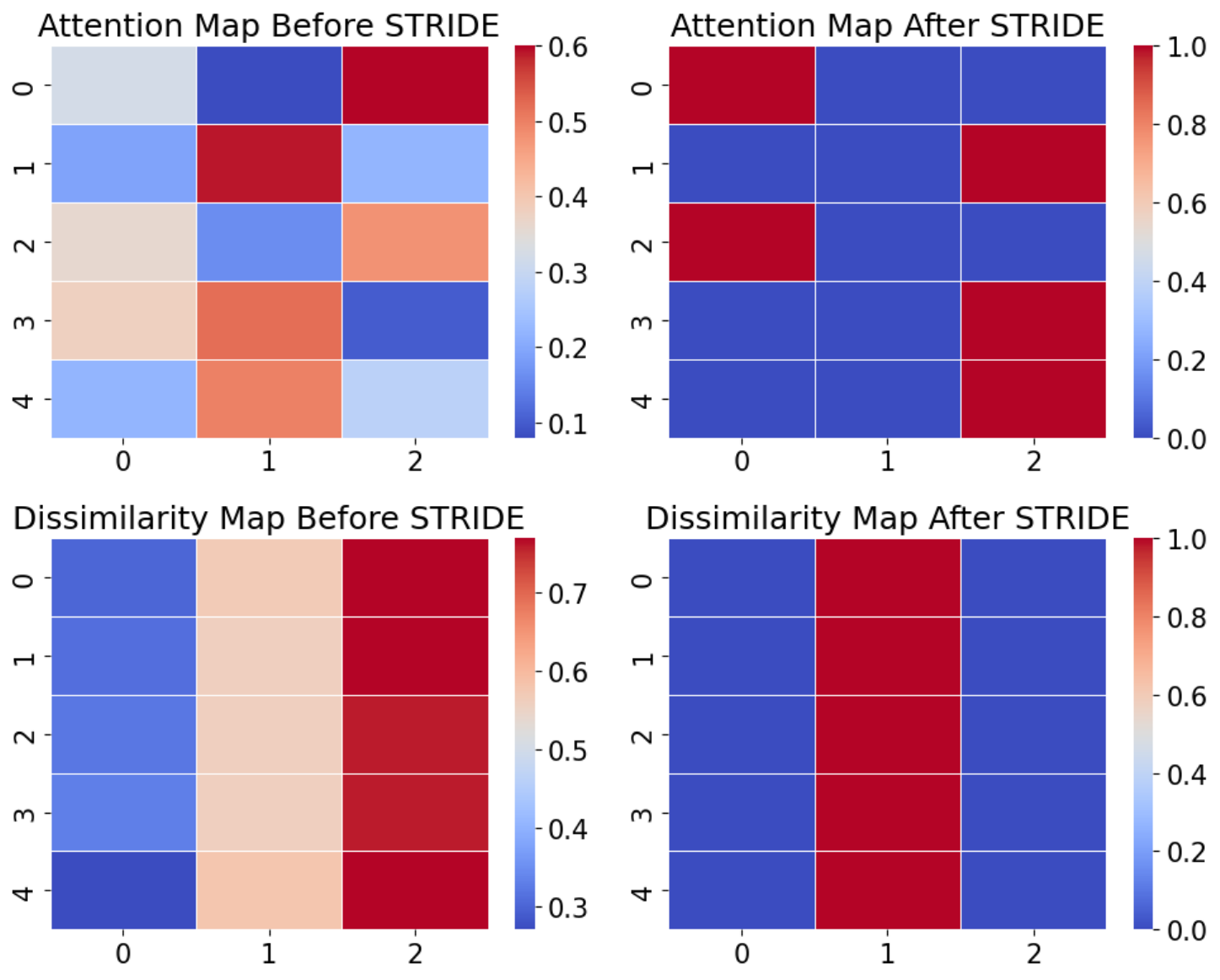}
\vspace{-2ex}
  \caption{Attention and Dissimilarity maps before and after training with \system. Cooler colors refer to lower scores and warmer colors correspond to higher scores.}
  \label{fig:att_maps_mag}
  \vspace{-2ex}
\end{figure}
\subsubsection{Aligning Intermediate Embeddings in\system} To empirically prove that \system aligns intermediate embeddings based on the attention matrix, we visualize the before and after training attention and dissimilarity maps in Figure~\ref{fig:att_maps_mag}. We train on OGBN-Mag and use a deeper teacher network of 5 layers and a hidden dimension of 512. The student network has 3 layers and a hidden dimension of 32. Our results show that dissimilarity scores are low where the attention scores are high and vice versa. This is in line with the intuition presented earlier in the \system mechanism. 
\begin{table}
    \centering
    \vspace{-2ex}
  \resizebox{0.48\textwidth}{!}{
            \begin{tabular}{r|cccc}
                \hline
                \textbf{Dataset} & \textbf{Cora} & \textbf{Citeseer} & \textbf{Pubmed} & \textbf{NELL} \\
                \textbf{Teacher} & GCNII (64L-64H) & GCNII (64L-64H) & GCNII (64L-64H) & GCNII (64L-64H) \\
                \textbf{Student} & GCNII (4L-4H) & GCNII (4L-4H) & GCNII (4L-4H) & GCNII (4L-4H) \\
                \hline
                Teacher & $88.40$ & $77.33$ & $89.78$ & $95.55$ \\
                Student & $73.87 \pm 0.42$ & $68.32 \pm 0.45$ & $87.87 \pm 0.45$ & $85.00 \pm 0.65$ \\
                LSP & $75.07 \pm 0.55 $ & $70.23 \pm 0.32$ & $88.07 \pm 0.45$ & $85.15 \pm 0.47$ \\
                GSP & $78.22 \pm 0.31$ & $69.50 \pm 0.67$ & $89.19 \pm 0.55$ & $86.32 \pm 0.45$ \\
                G-CRD & $83.45 \pm 0.45$ & $71.07 \pm 0.41$ & $89.66 \pm 0.48$ & $88.42 \pm 0.53$ \\ 
                \system & $\mathbf{84.27 \pm 0.32}$ & $\mathbf{72.00 \pm 0.30}$ & $\mathbf{89.89 \pm 0.31}$ & $\mathbf{92.02 \pm 0.64}$\\
                \hline
                \# St. Params & $5835$ & $14910$ & $2083$ & $22686$\\
                \hline
                Ratio & $60.7\times$ & $33.5\times$ & $141.3\times$ & $27.7\times $\\
                \hline
            \end{tabular}
\textbf{}        }
        \vspace{-2ex}
        \caption{Average accuracies for a variety of relatively smaller datasets. Each distillation method is applied to the same set of student weights. 
        %\payman{Why 40?}
        } \label{table:avgaccdeep-main}
        \vspace{-2ex}
\end{table}

\noindent \textbf{Deep GNNs}: We also test \system on deep GNN architectures (e.g., GCNII \cite{Chen_2020}). 
We test on Cora \cite{Mccallum_2000}, Citeseer \cite{Namata_2008}, Pubmed \cite{Namata_2012} and NELL \cite{Betteridge_2010}. Table~\ref{table:avgaccdeep-main} shows that \system can distill these deep GCNIIs into shallower GCNIIs with higher accuracy compared to other distillation methods. With over $27\times$ compression, \system achieves a $3.5\%$ accuracy improvement. Even at a $141\times$ compression ratio, \system matches the original teacher model's accuracy.

\subsubsection{Improved Weight Initialization for Highly Compressed Networks} 
\begin{figure}[!h]
   \centering
   \vspace{-2ex}
   \includegraphics[width=0.90\columnwidth]{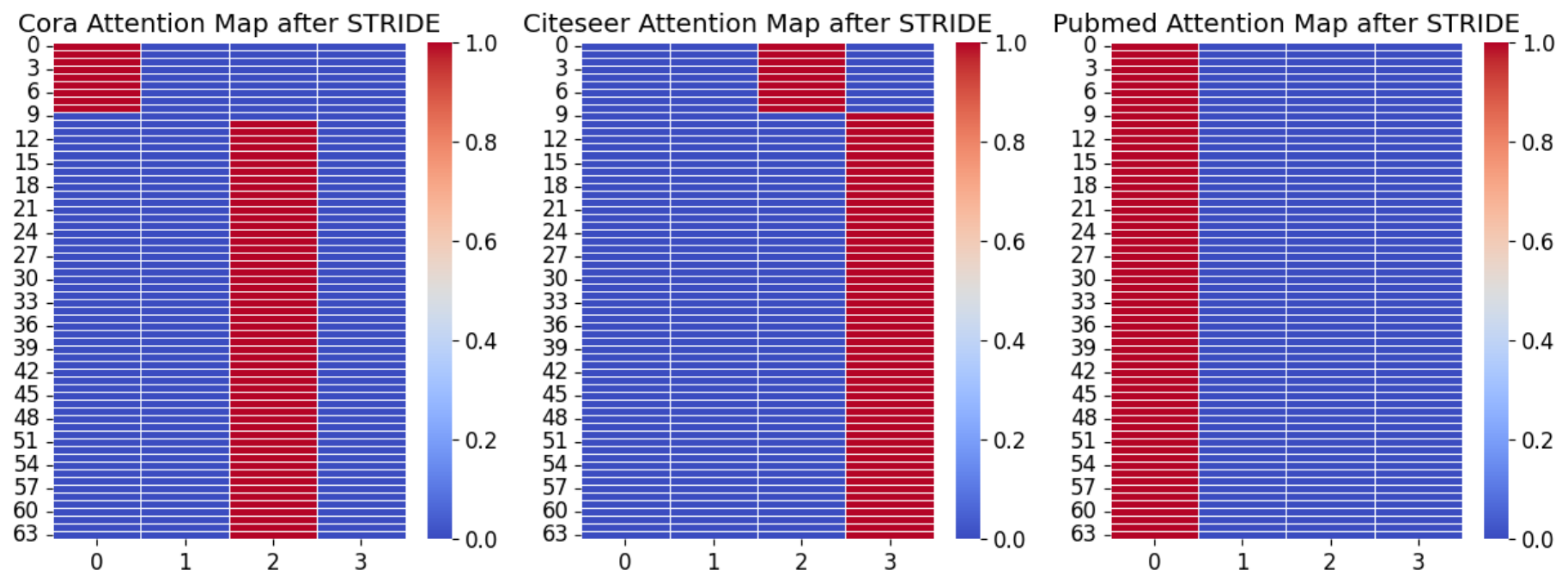}
   \vspace{-2ex}
  \caption{ Attention maps for Cora, Citeseer, and Pubmed. Each color in the heatmap represents the importance score associated with that teacher-student layer pair. Warmer colors mean higher importance scores. It seems apparent that most of the knowledge from the teacher layers is distilled into one student layer.}
  \label{fig:heatmap}
  \vspace{-2ex}
\end{figure}

We find that for smaller datasets, information from the teacher network is mainly distilled into one layer of the student network, as shown in Figure~\ref{fig:heatmap}. We hypothesize that smaller datasets lack complexity, allowing a single layer to capture most patterns.

To test this hypothesis, we first apply \system to a student network of arbitrary size and then generate the  attention map, $A \in R^{T_l \times S_l}$. 
The next step is to use a row-wise $argmax$ and find the student layer that has the most information distilled down to it. For example, in Figure~\ref{fig:heatmap}, the selected layer for Cora would be the third student layer (index 2 in the Figure). We then instantiate a new one-layer network and copy over the weights from the identified layer (as indicated by the attention map  $A$). 

We then evaluate this new network on the test set and report the results in column 3 of Table~\ref{table:wi1layer}. The first column of Table~\ref{table:wi1layer} represents the accuracies that we obtain after we train the new one-layer network for $1200$ epochs; we compare this result to the accuracy obtained from training a one-layer network from random initialization, which we report in the second column of Table~\ref{table:wi1layer}.
\vspace{-10pt}

\begin{figure}[!h]
    \centering
        \centering
        \resizebox{0.4\textwidth}{!}{
            \begin{tabular}{rccc}
                \toprule
                Dataset & Initialized & Random Init & No Training \\
                \midrule
                Cora & $80.35$ & $73.59$ & $65.36$ \\
                Citeseer & $70.21$ & $68.15$ & $54.20$ \\
                Pubmed & $88.80$ & $85.98$ & $72.90$ \\
                \bottomrule
            \end{tabular}
        }
        \vspace{-2ex}
         \captionof{table}{Results for weight initialization experiment. These are all one-layer networks.}\label{table:wi1layer}
         \vspace{-2ex}
\end{figure}   
\vspace{-2ex}

\subsection{Ablation Studies}
\label{sec:experiments:ablations}
We conduct a series of ablation studies in this subsection to further validate the effectiveness of \system; additional results are provided in the Appendix.
\subsubsection{Boosting Performance by Aligning both Structure and Embeddings} \system is novel partly because it aligns both graph structure and node embeddings across teacher and student networks. To demonstrate the advantage of aligning both structure and embeddings, we compare \system\ to variants that align only structure (S-\system) or only embeddings (E-\system). Results in Table~\ref{table:cS-S-E} show that aligning both structures and embeddings is better than aligning just one of them. 

\begin{figure}[!h]    
        \centering
        \resizebox{0.4\textwidth}{!}{
            \begin{tabular}{rcc}
                \toprule
                Dataset & OGBN-Mag &  OGBN-Arxiv \\
                \midrule
                \system & $\mathbf{47.55 \pm 0.28}$ & $\mathbf{73.67 \pm 0.49}$ \\
                E-\system & $46.02 \pm 0.48$ & $71.32 \pm 0.53$ \\
                S-\system & $45.82 \pm 0.60$ & $71.48 \pm 0.57$  \\
                \bottomrule
            \end{tabular}
        }
        \vspace{-2ex}
         \captionof{table}{\system vs. S-\system vs E-\system. Teacher/student model configuration are in Table~\ref{table:avgacc-main}.}\label{table:cS-S-E}
    \vspace{-2ex}
\end{figure}

\subsubsection{Importance of Aligning Intermediate Layers} To prove that aligning intermediate layers is necessary for superior performance, we experiment with a variant of \system, which we call modified \system, where we set $A \in R^{T_l \times S_l} $ to all zeros, but we set the bottom right value to 1. This indicates that we are only interested in the dissimilarity between the last layer node embeddings of the teacher and student models. The results in Table~\ref{table:cS-S-E} prove that we gain accuracy by considering the outputs of intermediate layers for both teacher and student models. In this experiment, we start from the same set of initialized weights for both the \system and modified \system approaches. \\
\begin{table}[h]
\vspace{-3ex}
    \centering
    \begin{tabular}{rcc}
        \toprule
        Dataset & OGBN-Mag &  OGBN-Arxiv \\
        \midrule
        Student & $44.46 \pm 0.54$ & $71.27 \pm 0.48$ \\
        Modified \system & $46.75 \pm 0.58$ & $71.76 \pm 0.52$ \\
        \system & $\mathbf{47.58 \pm 0.31}$ & $\mathbf{73.59 \pm 0.45}$  \\
        \bottomrule
    \end{tabular}
     \vspace{-2ex}
        \caption{Comparing modified \system that only considers aligning the last layer node embeddings with \system that considers intermediate layer node embeddings. Teacher/student model configurations are in Table~\ref{table:avgacc-main}.}\label{table:modifiedstraide-ab}
    \vspace{-2ex}
\end{table}
\subsubsection{Improvements due to Subspace Projection} In the ``Proposed Approach'' section, we introduced the concept of subspace projection as a way to alleviate issues caused by high-dimensional embedding spaces. While it is not needed for \system to work, as Table~\ref{table:subproj-ab} shows, it improves the results as the learned subspace projection matrix tends to be of lower rank than the embedding dimension. This indicates that we can project our feature maps into subspaces smaller than $R^{d_a}$, which increases the semantic value of the dissimilarity scores.

\vspace{-2ex}
\begin{figure}[h]
    %\vspace{-25pt}
    \centering
        \centering
        \resizebox{0.4\textwidth}{!}{
            \begin{tabular}{rcc}
                \toprule
                Dataset & Subspace Projection & No Projection \\
                \midrule
                Cora & $84.29 \pm 0.28$ & $83.78 \pm 0.35$\\
                Citeseer & $72.05 \pm 0.35$ & $71.34 \pm 0.41$\\
                Pubmed & $89.88 \pm 0.43$ & $88.76 \pm 0.53$ \\
                NELL & $91.95 \pm 0.58$ & $91.02 \pm 0.63$ \\
                OGBN-Mag & $47.54 \pm 0.32$ & $46.75 \pm 0.44$ \\
                OGBN-Arxiv & $73.58 \pm 0.50$ & $73.00 \pm 0.53$ \\
                \bottomrule
            \end{tabular}
        }
        \vspace{-2ex}
         \captionof{table}{Subspace projection impact. Teacher and student networks are the same as the ones in Tables~\ref{table:avgacc-main} and ~\ref{table:avgaccdeep-main}.}\label{table:subproj-ab}
         \vspace{-3ex}
\end{figure}

\subsubsection{Necessity of a Linear Layer for Each Hidden Layer} In our approach, we mentioned that each hidden teacher and student layer is assigned a linear layer for projection into $d_a$. This is because each layer represents its own $k$-hop neighborhood, and using just one linear layer would prove inadequate in capturing the full spectrum of essential semantic information contained within each layer. We run an experiment in which we use only one linear layer for the teacher and student projections. As Table ~\ref{table:linlayer-ab} demonstrates, there is an accuracy drop compared to the situation where we use individual linear layers for the projection.

\vspace{-2ex}
\begin{figure}[!h]    
        \centering
        \resizebox{0.48\textwidth}{!}{
            \begin{tabular}{rcc}
                \toprule
                Dataset & Multiple Linear Layers & One Linear Layer \\
                \midrule
                Cora & $84.29 \pm 0.28$ & $76.32 \pm 0.81$\\
                Citeseer & $72.05 \pm 0.35$ & $69.00 \pm 1.07$\\
                Pubmed & $89.88 \pm 0.43$ & $87.85 \pm 0.84$ \\
                NELL & $91.95 \pm 0.58$ & $85.67 \pm 0.72$ \\
                OGBN-Mag & $47.54 \pm 0.32$ & $45.02 \pm 0.67$ \\
                OGBN-Arxiv & $73.58 \pm 0.50$ & $70.98 \pm 0.79$ \\
                \bottomrule
            \end{tabular}
        }
        \vspace{-2ex}
        \captionof{table}{Linear layer per hidden layer affect. The teacher and student networks are the same as the ones described in Tables~\ref{table:avgacc-main} and ~\ref{table:avgaccdeep-main}.}\label{table:linlayer-ab}
        \vspace{-2ex}
\end{figure}

\section{Related Work}
\label{sec:related}
\textbf{KD for GNNs without Attention:} KD for GNNs is a relatively niche field that has been expanded recently. In LSP \cite{lspyang2021distilling}, the authors attempt to align node embeddings between the student and teacher networks by maximizing the similarity between embeddings that share edges. Since only node embeddings between connected edges are aligned, this KD method  preserves only local topology. Joshi et. al \cite{globalkdJoshi_2022} extend LSP and propose two different KD algorithms: Global Structure Preserving Distillation (GSP) and Global Contrastive Representation Distillation (G-CRD). GSP extends LSP  by considering all pairwise similarities among node features, not just pairwise similarities between nodes connected by edges. G-CRD implicitly preserves global topology by aligning the student and teacher node feature vectors via contrastive learning~\cite{contrastivelearning_2018}. These works are examples of methods that focus on aligning structure as they use relationships between different nodes to transfer knowledge from the teacher to the student.

Mustad \cite{kim2021compressing} distills a large teacher GNN into a one-layer student GNN by minimizing a distance function between the student's final node embeddings and the teacher's final node embeddings. Some studies use adversarial training methods to distill knowledge from a teacher to a student network. GraphAKD \cite{he2022compressing} treats the student network as a generator and trains a discriminator to distinguish between the final node embeddings of the student and teacher networks. Online Adversarial Distillation (OAD)~\cite{wang2021online} trains multiple student models and trains a discriminator to distinguish between the outputs of different student models. More recent approaches such as T2-GNN~\cite{huo2023t2}, KDGCL~\cite{wang2024lightweight}, and SA-MLP~\cite{chen2022sa}, further advocate for utilizing embedding features of GNNs to improve KD in GNNs.

\noindent \textbf{Adapting CNN-based KD Approaches to GNNs:} There have been several KD approaches that have been applied to CNNs that the GNN community has tried to adapt to GNNs, including Fitnets \cite{fitnetsromero2015} and Attention Transfer (AT)~ \cite{AT_2016}. These methods both compute a distance metric, such as mean-squared error between the last layer node embeddings of the student and teacher networks, and do not take into account the adjacency matrix; therefore, these approaches can all be categorized as aligning only embeddings. Using attention to find similarities across student and teacher layers is a concept explored in CNNs \cite{cnnabkdji2021show}. However, the ideas from this work cannot be applied to GNNs because the feature-comparison operations are not applicable to graph data. GNNs need special consideration in this regard compared to CNNs due to the non-spatial and unstructured form of graph data.

\noindent \textbf{KD for GNNs via Attention:} Several works have constructed an attention mechanism for KD in GNNs; however, these approaches focus on distilling knowledge from multiple teachers to a single student. MSKD \cite{Zhang_2022} uses an attention mechanism to assign weights to teacher networks in proportion to how much knowledge they should transfer to student networks. MulDE \cite{MulDE_2021} focuses on link prediction for knowledge graphs and uses a contrast attention mechanism to weigh soft labels from different teachers.

It is important to note that the above works only consider the node embeddings at the final layer of the teacher and student networks and aim to align them with one another in various ways. GeometricKD \cite{geometricyang2023} aligns all teacher and student node embeddings, but it constrains the student and teacher networks to have the same number of layers to enforce a 1-1 correspondence between teacher and student layers. It then proceeds to align teacher layer $i$ with student layer $i$; this approach is inflexible as it severely constrains the student architecture.

Table~\ref{table:1} summarizes the main features of closely related work and how they are different from \system. Unlike existing works, \system aligns both structure and embeddings across all layers, without requiring strict architectural matching between teacher and student. This enables a richer transfer of hierarchical graph information and makes the approach applicable across diverse teacher-student architectures.

\begin{table}[!ht]
\centering
\vspace{-1ex}
\resizebox{0.47\textwidth}{!}{
\begin{tabular}{|c|c|c|c|}
\hline
Method  & Aligns Structure & Aligns Embeddings & Number of Layers Considered \\
\hline
GraphAKD & $\times$ & \checkmark & $1$  \\
G-CRD & \checkmark & $\times$ & $1$ \\
LSP    & \checkmark &  $\times$ &  $1$    \\
GSP    & \checkmark &  $\times$ &  $1$    \\
Fitnets & $\times$ & \checkmark & $1$   \\
AT    & $\times$ & \checkmark  &  $1$    \\
\system & \checkmark & \checkmark & \textbf{All} \\
\hline
\end{tabular}
}
\vspace{-2ex}
\caption{Comparison of various KD approaches with \system
}\label{table:1}
\end{table}

\vspace{-3ex}
\section{Conclusion}
\label{sec:conclusion}
The ever-growing size and complexity of GNNs pose problems such as increased computational complexity, memory usage, storage requirements, over-smoothing, and over-squashing, which complicate their practical deployment for various applications such as real-time recommendation systems, spam detection, and rapid image retrieval. To address this difficulty, we propose an innovative solution known as Structure and Embedding Distillation with Attention (\system). \system employs an attention-based feature linking mechanism to identify important intermediate teacher-student layer pairs and focuses on aligning the node embeddings and graph structure of those pairs. This KD approach broadly outperforms existing KD approaches for GNNs over a wide variety of compression settings. It also works with both deep and shallow networks, and shows robust performance with different GNN architectures. On average, we achieve a $2.13\%$ increase in accuracy with a $32.3\times$ compression ratio on OGBN-Mag, a large graph dataset, compared to state-of-the-art approaches. On smaller datasets (e.g., Pubmed), STRIDE matches the accuracy of state-of-the-art methods while achieving a $141\times$ compression ratio.

%\bibliography{main}

\begin{thebibliography}{55}
\providecommand{\natexlab}[1]{#1}

\bibitem[{Airas and Zhang(2025)}]{airas2025scaling}
Airas, J.; and Zhang, B. 2025.
\newblock Scaling Graph Neural Networks to Large Proteins.
\newblock \emph{Journal of chemical theory and computation}, 21(4): 2055--2066.

\bibitem[{Baxter(2000)}]{Baxter_2000}
Baxter, J. 2000.
\newblock A Model of Inductive Bias Learning.
\newblock \emph{Journal of Artificial Intelligence Research (JAIR)}, 12: 149--198.

\bibitem[{Carlson et~al.(2010)Carlson, Betteridge, Kisiel, Settles, Hruschka, and Mitchell}]{Betteridge_2010}
Carlson, A.; Betteridge, J.; Kisiel, B.; Settles, B.; Hruschka, E.; and Mitchell, T. 2010.
\newblock Toward an Architecture for Never-Ending Language Learning.
\newblock \emph{Proceedings of the AAAI Conference on Artificial Intelligence (AAAI)}, 1306--1313.

\bibitem[{Chen et~al.(2024{\natexlab{a}})Chen, Bai, Chen, Gao, Zhang, and Pu}]{chen2022sa}
Chen, J.; Bai, M.; Chen, S.; Gao, J.; Zhang, J.; and Pu, J. 2024{\natexlab{a}}.
\newblock SA-MLP: Distilling Graph Knowledge from GNNs into Structure-Aware MLP.
\newblock \emph{Transactions on Machine Learning Research (TMLR)}.

\bibitem[{Chen et~al.(2020)Chen, Wei, Huang, Ding, and Li}]{Chen_2020}
Chen, M.; Wei, Z.; Huang, Z.; Ding, B.; and Li, Y. 2020.
\newblock Simple and Deep Graph Convolutional Networks.
\newblock In \emph{International Conference on Machine Learning (ICML)}, 1725--1735.

\bibitem[{Chen et~al.(2024{\natexlab{b}})Chen, He, Qu, and Zhang}]{chen2024hopgnn}
Chen, W.; He, S.; Qu, H.; and Zhang, X. 2024{\natexlab{b}}.
\newblock HopGNN: Boosting Distributed GNN Training Efficiency via Feature-Centric Model Migration.
\newblock \emph{Preprint arXiv:2409.00657}.

\bibitem[{Di~Giovanni et~al.(2023)Di~Giovanni, Giusti, Barbero, Luise, Lio, and Bronstein}]{di2023over}
Di~Giovanni, F.; Giusti, L.; Barbero, F.; Luise, G.; Lio, P.; and Bronstein, M.~M. 2023.
\newblock On Over-Squashing in Message Passing Neural Networks: The Impact of Width, Depth, and Topology.
\newblock In \emph{International Conference on Machine Learning (ICML)}, 7865--7885.

\bibitem[{Esfahani et~al.(2023)Esfahani, Boldi, Vandierendonck, Kilpatrick, and Vigna}]{esfahani2023ms}
Esfahani, M.~K.; Boldi, P.; Vandierendonck, H.; Kilpatrick, P.; and Vigna, S. 2023.
\newblock MS-BioGraphs: Sequence Similarity Graph Datasets.
\newblock \emph{Preprint arXiv:2308.16744}.

\bibitem[{Formal et~al.(2020)Formal, Clinchant, Renders, Lee, and Cho}]{Formal_2020}
Formal, T.; Clinchant, S.; Renders, J.-M.; Lee, S.; and Cho, G.~H. 2020.
\newblock Learning to Rank Images with Cross-Modal Graph Convolutions.
\newblock In \emph{the European Conference on Information Retrieval (ECIR)}, 589--604.

\bibitem[{{GraphGeeks Lab}(2024)}]{graphexplorer}
{GraphGeeks Lab}. 2024.
\newblock {Graph Explorer}.
\newblock \url{https://github.com/graphgeeks-lab/graph-explorer}.
\newblock [Online; accessed December-2024].

\bibitem[{Hamilton, Ying, and Leskovec(2017)}]{GraphSAGE_2017}
Hamilton, W.~L.; Ying, Z.; and Leskovec, J. 2017.
\newblock Inductive Representation Learning on Large Graphs.
\newblock In \emph{Advances in Neural Information Processing Systems (NeurIPS)}, 1024--1034.

\bibitem[{He et~al.(2022)He, Wang, Zhang, and Wu}]{he2022compressing}
He, H.; Wang, J.; Zhang, Z.; and Wu, F. 2022.
\newblock Compressing Deep Graph Neural Networks via Adversarial Knowledge Distillation.
\newblock In \emph{Proceedings of the 28th ACM SIGKDD conference on Knowledge Discovery and Data mining (KDD)}, 534--544.

\bibitem[{Hinton, Vinyals, and Dean(2014)}]{hinton2015distilling}
Hinton, G.; Vinyals, O.; and Dean, J. 2014.
\newblock Distilling the Knowledge in a Neural Network.
\newblock In \emph{Proceedings of the NeurIPS 2014 Deep Learning Workshop (NeurIPS Workshop)}.

\bibitem[{Hu et~al.(2020)Hu, Fey, Zitnik, Dong, Ren, Liu, Catasta, and Leskovec}]{OGBN_2020}
Hu, W.; Fey, M.; Zitnik, M.; Dong, Y.; Ren, H.; Liu, B.; Catasta, M.; and Leskovec, J. 2020.
\newblock Open Graph Benchmark: Datasets for Machine Learning on Graphs.
\newblock In \emph{Advances in Neural Information Processing Systems (NeurIPS)}, 22118--22133.

\bibitem[{Huang et~al.(2021)Huang, Zhang, Xi, Liu, and Zhou}]{huang2021scaling}
Huang, Z.; Zhang, S.; Xi, C.; Liu, T.; and Zhou, M. 2021.
\newblock Scaling Up Graph Neural Networks via Graph Coarsening.
\newblock In \emph{Proceedings of the 27th ACM SIGKDD conference on Knowledge Discovery \& Data mining (KDD)}, 675--684.

\bibitem[{Huo et~al.(2023)Huo, Jin, Li, He, Yang, and Wu}]{huo2023t2}
Huo, C.; Jin, D.; Li, Y.; He, D.; Yang, Y.-B.; and Wu, L. 2023.
\newblock T2-GNN: Graph Neural Networks for Graphs with Incomplete Features and Structure via Teacher-Student Distillation.
\newblock In \emph{Proceedings of the AAAI Conference on Artificial Intelligence (AAAI)}, 4339--4346.

\bibitem[{Ji, Heo, and Park(2021)}]{cnnabkdji2021show}
Ji, M.; Heo, B.; and Park, S. 2021.
\newblock Show, Attend and Distill: Knowledge Distillation via Attention-based Feature Matching.
\newblock In \emph{Proceedings of the AAAI Conference on Artificial Intelligence (AAAI)}, 7945--7952.

\bibitem[{Jing et~al.(2021)Jing, Yang, Wang, Song, and Tao}]{amalgamateYongcheng2021cvpr}
Jing, Y.; Yang, Y.; Wang, X.; Song, M.; and Tao, D. 2021.
\newblock Amalgamating Knowledge from Heterogeneous Graph Neural Networks.
\newblock In \emph{Proceedings of the IEEE/CVF Conference on Computer Vision and Pattern Recognition (CVPR)}, 15704--15713.

\bibitem[{Joshi et~al.(2022)Joshi, Liu, Xun, Lin, and Foo}]{globalkdJoshi_2022}
Joshi, C.~K.; Liu, F.; Xun, X.; Lin, J.; and Foo, C.~S. 2022.
\newblock On Representation Knowledge Distillation for Graph Neural Networks.
\newblock \emph{{IEEE} Transactions on Neural Networks and Learning Systems (TNNLS)}, 1--12.

\bibitem[{Kim, Jung, and Kang(2021)}]{kim2021compressing}
Kim, J.; Jung, J.; and Kang, U. 2021.
\newblock Compressing Deep Graph Convolution Network with Multi-Staged Knowledge Distillation.
\newblock \emph{PLoS ONE}, 16(8).

\bibitem[{Kiningham, Levis, and R{\'e}(2022)}]{kiningham2022grip}
Kiningham, K.; Levis, P.; and R{\'e}, C. 2022.
\newblock GRIP: A Graph Neural Network Accelerator Architecture.
\newblock \emph{IEEE Transactions on Computers (TC)}, 72(4): 914--925.

\bibitem[{Kipf and Welling(2017)}]{KipfW16}
Kipf, T.~N.; and Welling, M. 2017.
\newblock Semi-Supervised Classification with Graph Convolutional Networks.
\newblock \emph{The International Conference on Learning Representations (ICLR)}.

\bibitem[{Li et~al.(2019)Li, Qin, Liu, Yang, and Li}]{Li_2019}
Li, A.; Qin, Z.; Liu, R.; Yang, Y.; and Li, D. 2019.
\newblock Spam Review Detection with Graph Convolutional Networks.
\newblock In \emph{Proceedings of the 28th ACM International Conference on Information and Knowledge Management (CIKM)}, 2703–2711.

\bibitem[{Li et~al.(2021)Li, Yuan, Radfar, Marendy, Ni, O’Brien, and Casillas-Espinosa}]{li2021graph}
Li, R.; Yuan, X.; Radfar, M.; Marendy, P.; Ni, W.; O’Brien, T.~J.; and Casillas-Espinosa, P.~M. 2021.
\newblock Graph signal processing, Graph Neural Network and Graph Learning on Biological Data: a Systematic Review.
\newblock \emph{IEEE Reviews in Biomedical Engineering}, 16: 109--135.

\bibitem[{Liu et~al.(2024{\natexlab{a}})Liu, Hooi, Kawaguchi, Wang, Dong, and Xiao}]{liu2024scalable}
Liu, J.; Hooi, B.; Kawaguchi, K.; Wang, Y.; Dong, C.; and Xiao, X. 2024{\natexlab{a}}.
\newblock Scalable and Effective Implicit Graph Neural Networks on Large Graphs.
\newblock In \emph{The International Conference on Learning Representations (ICLR)}.

\bibitem[{Liu et~al.(2024{\natexlab{b}})Liu, Mao, Chen, Zhao, Shah, and Tang}]{liu2024neural}
Liu, J.; Mao, H.; Chen, Z.; Zhao, T.; Shah, N.; and Tang, J. 2024{\natexlab{b}}.
\newblock Neural Scaling Laws on Graphs.
\newblock \emph{Preprint arXiv:2402.02054}.

\bibitem[{Liu et~al.(2022)Liu, Zou, Zou, Wang, Zhang, Tang, Zhu, Zhu, Wu, Wang, and Cheng}]{liu_2022}
Liu, Z.; Zou, L.; Zou, X.; Wang, C.; Zhang, B.; Tang, D.; Zhu, B.; Zhu, Y.; Wu, P.; Wang, K.; and Cheng, Y. 2022.
\newblock Monolith: Real Time Recommendation System With Collisionless Embedding Table.
\newblock In \emph{Proceedings of the 5th Workshop on Online Recommender Systems and User Modeling (ORSUM), in conjunction with the 16th {ACM} Conference on Recommender Systems}.

\bibitem[{Ma, Rong, and Huang(2022)}]{ma2022graph}
Ma, H.; Rong, Y.; and Huang, J. 2022.
\newblock Graph Neural Networks: Scalability.
\newblock \emph{Graph Neural Networks: Foundations, Frontiers, and Applications}, 99--119.

\bibitem[{Mccallum, Nigam, and Rennie(2000)}]{Mccallum_2000}
Mccallum, A.; Nigam, K.; and Rennie, J. 2000.
\newblock Automating the Construction of Internet Portals.
\newblock \emph{Information Retrieval}, 3: 127--163.

\bibitem[{Namata et~al.(2012)Namata, London, Getoor, and Huang}]{Namata_2012}
Namata, G.; London, B.; Getoor, L.; and Huang, B. 2012.
\newblock Query-driven Active Surveying for Collective Classification.
\newblock In \emph{International Workshop on Mining and Learning with graphs (MLG)}.

\bibitem[{Oord, Li, and Vinyals(2018)}]{contrastivelearning_2018}
Oord, A. v.~d.; Li, Y.; and Vinyals, O. 2018.
\newblock Representation Learning with Contrastive Predictive Coding.
\newblock \emph{Preprint arXiv:1807.03748}.

\bibitem[{Que et~al.(2024)Que, Fan, Loo, Li, Blott, Pierini, Tapper, and Luk}]{que_2023}
Que, Z.; Fan, H.; Loo, M.; Li, H.; Blott, M.; Pierini, M.; Tapper, A.; and Luk, W. 2024.
\newblock LL-GNN: Low Latency Graph Neural Networks on FPGAs for High Energy Physics.
\newblock \emph{ACM Transactions on Embedded Computing Systems (TECS)}, 23(2): 1--28.

\bibitem[{Romero et~al.(2015)Romero, Ballas, Ebrahimi~Kahou, Chassang, Gatta, and Bengio}]{fitnetsromero2015}
Romero, A.; Ballas, N.; Ebrahimi~Kahou, S.; Chassang, A.; Gatta, C.; and Bengio, Y. 2015.
\newblock FitNets: Hints for Thin Deep Nets.
\newblock In \emph{International Conference on Learning Representations (ICLR)}.

\bibitem[{Sarkar et~al.(2023)Sarkar, Abi-Karam, He, Sathidevi, and Hao}]{sarkar2023flowgnn}
Sarkar, R.; Abi-Karam, S.; He, Y.; Sathidevi, L.; and Hao, C. 2023.
\newblock FlowGNN: A Dataflow Architecture for Real-time Workload-agnostic Graph Neural Network Inference.
\newblock In \emph{2023 IEEE International Symposium on High-Performance Computer Architecture (HPCA)}, 1099--1112.

\bibitem[{Schlichtkrull et~al.(2018)Schlichtkrull, Kipf, Bloem, Van Den~Berg, Titov, and Welling}]{RGCN_2017}
Schlichtkrull, M.; Kipf, T.~N.; Bloem, P.; Van Den~Berg, R.; Titov, I.; and Welling, M. 2018.
\newblock Modeling Relational Data with Graph Convolutional Networks.
\newblock In \emph{European Semantic Web Conference (ESWC)}, 593--607.

\bibitem[{Sen et~al.(2008)Sen, Namata, Bilgic, Getoor, Galligher, and Eliassi-Rad}]{Namata_2008}
Sen, P.; Namata, G.; Bilgic, M.; Getoor, L.; Galligher, B.; and Eliassi-Rad, T. 2008.
\newblock Collective Classification in Network Data.
\newblock \emph{AI Magazine}, 29: 93.

\bibitem[{Shi and Rajkumar(2020)}]{shi2020point}
Shi, W.; and Rajkumar, R. 2020.
\newblock Point-GNN: Graph Neural Network for 3D Object Detection in a Point Cloud.
\newblock In \emph{Proceedings of the IEEE/CVF conference on Computer Vision and Pattern Recognition (CVPR)}, 1711--1719.

\bibitem[{Shlomi, Battaglia, and Vlimant(2020)}]{shlomi2020graph}
Shlomi, J.; Battaglia, P.; and Vlimant, J.-R. 2020.
\newblock Graph Neural Networks in Particle Physics.
\newblock \emph{Machine Learning: Science and Technology (MLST)}, 2(2).

\bibitem[{Sypetkowski et~al.(2023)Sypetkowski, Wenkel, Poursafaei, Dickson, Suri, Fradkin, and Beaini}]{sypetkowski2024scalability}
Sypetkowski, M.; Wenkel, F.; Poursafaei, F.; Dickson, N.; Suri, K.; Fradkin, P.; and Beaini, D. 2023.
\newblock On the Scalability of GNNs for Molecular Graphs.
\newblock In \emph{Advances in Neural Information Processing Systems (NeurIPS)}, 19870--19906.

\bibitem[{Tan et~al.(2023)Tan, Yuan, He, Sit, Li, Liu, Ai, Zeng, Pietzuch, and Mai}]{tan_2023}
Tan, Z.; Yuan, X.; He, C.; Sit, M.-K.; Li, G.; Liu, X.; Ai, B.; Zeng, K.; Pietzuch, P.; and Mai, L. 2023.
\newblock Quiver: Supporting GPUs for Low-Latency, High-throughput GNN Serving with Workload Awareness.
\newblock \emph{Preprint arXiv:2305.10863}.

\bibitem[{Tian et~al.(2025)Tian, Pei, Zhang, Zhang, and Chawla}]{tian2023knowledge}
Tian, Y.; Pei, S.; Zhang, X.; Zhang, C.; and Chawla, N.~V. 2025.
\newblock Knowledge Distillation on Graphs: A Survey.
\newblock \emph{ACM Computing Surveys}, 57(8): 1--16.

\bibitem[{Uselis and Oh(2025)}]{uselisintermediate-2025}
Uselis, A.; and Oh, S.~J. 2025.
\newblock Intermediate Layer Classifiers for OOD Generalization.
\newblock In \emph{The International Conference on Learning Representations (ICLR)}.

\bibitem[{Veli{\v{c}}kovi{\'c} et~al.(2018)Veli{\v{c}}kovi{\'c}, Cucurull, Casanova, Romero, Li{\`o}, and Bengio}]{veličković_2018}
Veli{\v{c}}kovi{\'c}, P.; Cucurull, G.; Casanova, A.; Romero, A.; Li{\`o}, P.; and Bengio, Y. 2018.
\newblock Graph Attention Networks.
\newblock In \emph{International Conference on Learning Representations (ICLR)}.

\bibitem[{Wang et~al.(2021{\natexlab{a}})Wang, Wang, Chen, Zhou, Feng, and Chen}]{wang2021online}
Wang, C.; Wang, Z.; Chen, D.; Zhou, S.; Feng, Y.; and Chen, C. 2021{\natexlab{a}}.
\newblock Online Adversarial Distillation for Graph Neural Networks.
\newblock \emph{Preprint arXiv:2112.13966}.

\bibitem[{Wang et~al.(2021{\natexlab{b}})Wang, Liu, Ma, and Sheng}]{MulDE_2021}
Wang, K.; Liu, Y.; Ma, Q.; and Sheng, Q.~Z. 2021{\natexlab{b}}.
\newblock MulDE: Multi-Teacher Knowledge Distillation for Low-Dimensional Knowledge Graph Embeddings.
\newblock In \emph{Proceedings of the Web Conference (WWW)}, 1716–1726.

\bibitem[{Wang et~al.(2020)Wang, Shen, Huang, Wu, Dong, and Kanakia}]{MAG_2020}
Wang, K.; Shen, Z.; Huang, C.; Wu, C.-H.; Dong, Y.; and Kanakia, A. 2020.
\newblock {Microsoft Academic Graph: When Experts Are not Enough}.
\newblock \emph{Quantitative Science Studies (QSS)}, 1(1): 396--413.

\bibitem[{Wang and Yang(2024)}]{wang2024lightweight}
Wang, Y.; and Yang, S. 2024.
\newblock A Lightweight Method for Graph Neural Networks Based on Knowledge Distillation and Graph Contrastive Learning.
\newblock \emph{Applied Sciences}, 14(11): 4805.

\bibitem[{{Web Data Commons}(2024)}]{hyperlinkgraph}
{Web Data Commons}. 2024.
\newblock {Web Data Commons Hyperlink Graph}.
\newblock \url{https://webdatacommons.org/hyperlinkgraph/}.
\newblock [Online; accessed December-2024].

\bibitem[{{Wikipedia contributors}(2024)}]{dbpedia}
{Wikipedia contributors}. 2024.
\newblock {DBpedia --- Wikipedia, The Free Encyclopedia}.
\newblock \url{https://en.wikipedia.org/wiki/DBpedia}.
\newblock [Online; accessed December-2024].

\bibitem[{Yang, Wu, and Yan(2022)}]{geometricyang2023}
Yang, C.; Wu, Q.; and Yan, J. 2022.
\newblock Geometric Knowledge Distillation: Topology Compression for Graph Neural Networks.
\newblock \emph{Advances in Neural Information Processing Systems (NeurIPS)}, 29761--29775.

\bibitem[{Yang et~al.(2020)Yang, Qiu, Song, Tao, and Wang}]{lspyang2021distilling}
Yang, Y.; Qiu, J.; Song, M.; Tao, D.; and Wang, X. 2020.
\newblock Distilling Knowledge from Graph Convolutional Networks.
\newblock In \emph{Proceedings of the IEEE/CVF conference on Computer Vision and Pattern Recognition (CVPR)}, 7074--7083.

\bibitem[{Yang et~al.(2022)Yang, Zhong, Zhao, and Chen}]{yang2022mgraphdta}
Yang, Z.; Zhong, W.; Zhao, L.; and Chen, C. Y.-C. 2022.
\newblock MGraphDTA: Deep Multiscale Graph Neural Network for Explainable Drug--Target Binding Affinity Prediction.
\newblock \emph{Chemical science}, 13(3): 816--833.

\bibitem[{Zagoruyko and Komodakis(2017)}]{AT_2016}
Zagoruyko, S.; and Komodakis, N. 2017.
\newblock Paying More Attention to Attention: Improving the Performance of Convolutional Neural Networks via Attention Transfer.
\newblock In \emph{International Conference on Learning Representations (ICLR)}.

\bibitem[{Zhang et~al.(2022)Zhang, Liu, Dang, and Zhang}]{Zhang_2022}
Zhang, C.; Liu, J.; Dang, K.; and Zhang, W. 2022.
\newblock Multi-scale Distillation from Multiple Graph Neural Networks.
\newblock In \emph{Proceedings of the AAAI Conference on Artificial Intelligence (AAAI)}, 4337--4344.

\bibitem[{Zhou et~al.(2021)Zhou, Srivastava, Zeng, Kannan, and Prasanna}]{zhou2021accelerating}
Zhou, H.; Srivastava, A.; Zeng, H.; Kannan, R.; and Prasanna, V. 2021.
\newblock Accelerating large scale real-time {GNN} inference using channel pruning.
\newblock \emph{Proceedings of the {VLDB} Endowment (VLDB)}, 14(9): 1597--1605.

\end{thebibliography}

\clearpage
\appendix
\section{Appendix}

\subsection{Latency Increase by Number of Parameters}
\begin{figure}[!h]
\centering
  \includegraphics[width=0.4\textwidth]{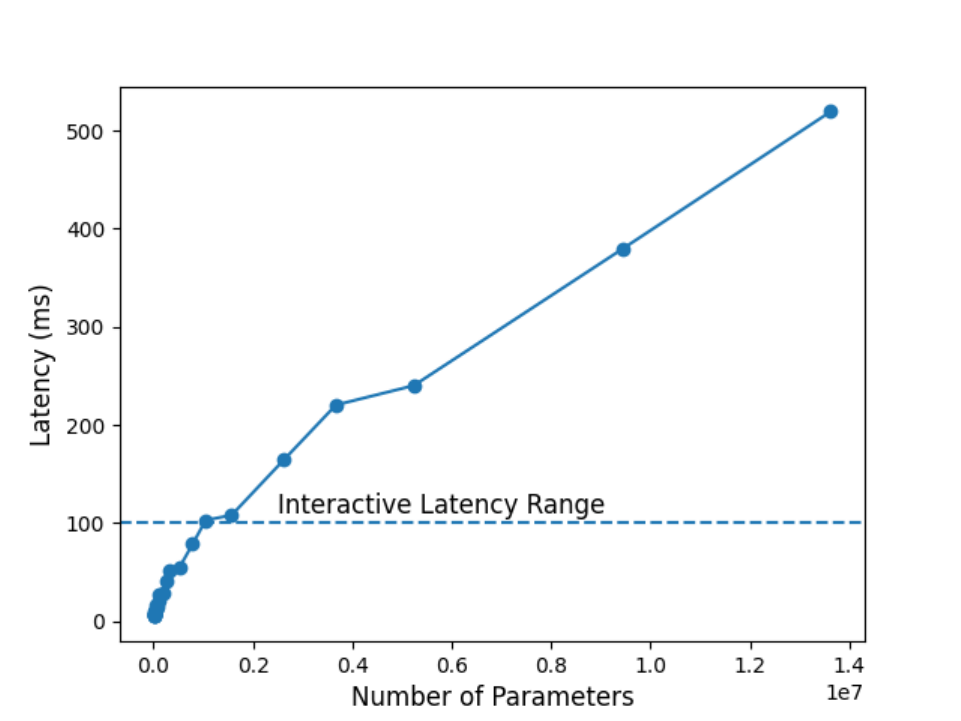}
  \vspace{-1ex}
\caption{\small  Inference latency of a standard GCN model architecture by increasing number of parameters (e.g., embedding dimension) on Flickr dataset.  All tests were run on a Tesla V100 GPU with a batch size of 1.}\label{fig:motivation}
\end{figure}

Figure~\ref{fig:motivation} demonstrates how increasing the number of model parameters directly impacts inference latency. To visualize this trend, we plot the inference latency of a standard GCN model as we scale up its parameter count (e.g., by enlarging the embedding dimension) on Flickr dataset. The figure reveals that as the model becomes more expressive and parameter-heavy, inference time rises substantially. .

\subsection{Dataset and Teacher Network Information}
\label{sec:appendix_b:dataset_info}
Table \ref{tab:spec} provides specifications on the datasets used in our experiments. 

\vspace{5pt}
\begin{table}[ht]
\centering
\resizebox{0.48\textwidth}{!}{
\vspace{-4ex}
\begin{tabular}{c|c|c|c|c}
\toprule
         & \# of Nodes & \# of Edges & \# of Features & \# of Classes \\
Cora     & 2,708       & 10,556      & 1,433          & 7             \\
Citeseer & 3,327       & 9,104       & 3,703          & 6             \\
Pubmed   & 19,717      & 88,648      & 500            & 3             \\
NELL   & 65,755      & 251,550     & 61,278            & 186 \\
Flickr   &89,250    & 899,756     & 500           & 7 \\
OGBN-Mag & 1,939,743   & 21,111,007  & 128            & 349           \\
OGBN-Arxiv & 169,343   & 1,166,243   & 128            & 40            \\
\bottomrule
\end{tabular}}
\vspace{-1ex}
\caption{Specification of evaluated datasets.}
\label{tab:spec}
\end{table}

\subsection{Ablation Studies on Hyperparameter} 
Two main hyperparameters need to be tuned when training \system: the loss coefficient, $\beta$, and the \system embedding dimension, $d_a$. We present the test accuracies across various values for $\beta$ and $d_a$ in Tables \ref{tab:ablationbeta} and \ref{tab:ablationda}. These results show that a relatively lower $\beta$ and a higher $d_a$ tend to produce slightly better results. For all of our experiments, we used a $\beta$ of $10$ and a $d_a$ of 256 for this reason.
\vspace{5pt}
\begin{table}[h]
    \centering
        \resizebox{0.48\textwidth}{!}{
            \begin{tabular}{rcccc}
                \toprule
                Dataset & $\beta=1$ & $\beta=10$ & $\beta=20$ & $\beta=50$ \\
                \midrule
                Cora & $86.92$ & $84.71$ & $86.19$ & $85.64$\\
                Citeseer & $73.20$ & $74.33$ & $71.82$ & $69.82$\\
                Pubmed & $89.03$ & $89.97$ & $88.32$ & $88.56$ \\
                NELL & $90.86$ & $88.73$ & $90.14$ & $91.32$ \\
                \bottomrule
            \end{tabular}
        }
        \vspace{-2ex}
        \captionof{table}{Ablation results for $\beta$.}
        \label{tab:ablationbeta}
\end{table}  

\begin{table}[h]
        \resizebox{0.48\textwidth}{!}{
            \begin{tabular}{rcccc}
                \toprule
                Dataset & $d_a=64$ & $d_a=128$ & $d_a=256$ & $d_a=512$ \\
                \midrule
                Cora & $87.45$ & $87.11$ & $86.92$ & $86.37$\\
                Citeseer & $72.07$ & $72.52$ & $73.12$ & $74.62$\\
                Pubmed & $89.58$ & $89.58$ & $89.33$ & $89.12$\\
                NELL & $89.73$ & $90.12$ & $90.73$ & $91.14$ \\
                \bottomrule
            \end{tabular}
        }
        \vspace{-1ex}
        \captionof{table}{Ablation results for $d_a$.}
        \label{tab:ablationda}
\end{table}

\subsection{Euclidean vs. Cosine Distance} 
In the ``Proposed'' section, we mention that we use an Euclidean distance metric instead of a cosine distance metric to generate the dissimilarity matrix, $M$. We present the results of this ablation in Table \ref{tab:EuclideanVScosine}. Cosine distance, which only considers the direction of vectors and not their magnitude, may disregard critical information contained in the magnitude of hidden representations. We hypothesize that preserving this information, as done by Euclidean distance, is important to its superior performance observed in our experiments. 

\begin{table}[h]
    \centering
    \begin{tabular}{rcc}
        \toprule
        Dataset & Euclidean Distance & Cosine Distance \\
        \midrule
        Cora & $87.33$ & $82.81$\\
        Citeseer & $73.43$ & $70.98$\\
        Pubmed & $89.58$ & $87.32$ \\
        NELL & $91.24$ & $85.66$ \\
        \bottomrule
    \end{tabular}
    \vspace{-1ex}
    \caption{Test accuracies when using Euclidean vs. cosine distance for computing dissimilarity matrix $M$.} 
    \label{tab:EuclideanVScosine}
\end{table}

\subsection{Details of the Theorem the and Proof}

\setcounter{theorem}{0}
\begin{theorem}[STRIDE Cross‑Layer Gradient Dependence]
Let 
\[
L_{\text{STRIDE}}
     = \mathbf 1_{T_\ell}^{\!\top}
       \!\bigl(\boldsymbol{\alpha}\odot\mathbf M\bigr)
       \frac{\mathbf 1_{S_\ell}}{S_\ell},
       \qquad
       \mathbf M=\mathbf D_{\text{emb}}+\mathbf D_{\text{str}}
       \tag{6}
\]
be the distillation loss defined in Eq.(6) of the main paper,  
where  
\(\boldsymbol{\alpha}\in\mathbb R^{T_\ell\times S_\ell}\) is the attention matrix of Eq.(3) and  
\(\mathbf M\) collects the pair‑wise embedding and structural dissimilarities of Eqs.(4)–(5).  
For every student layer \(j\in[1,S_\ell]\) the gradient of \(L_{\text{\em STRIDE}}\) with respect
to the student weight matrix \(W_j^{s}\) depends on \emph{every} teacher weight
matrix \(W_i^{t}\,(i=1,\dots,T_\ell)\):
\[
\boxed{
  \frac{\partial L_{\text{STRIDE}}}{\partial W_j^{s}}
   =f\!\bigl(\{W_i^{t}\}_{i=1}^{T_\ell},\{W_\ell^{s}\}_{\ell=1}^{S_\ell}\bigr)
  }\qquad\forall\,j .
\]
Consequently, gradients flow from \emph{all} teacher layers—even those deeper
than the student layer (\(i>j\))—directly into \(W_j^{s}\).
\end{theorem}

\vspace{6ex}
\textbf{Proof:}
For a single teacher–student layer pair \((i,j)\) define
\[
L_{ij}\;:=\;\alpha_{ij}\,M_{ij},\qquad
\alpha_{ij}=\frac{\exp z_{ij}}{\sum_{i',j'}\exp z_{i'j'}},
\tag{A1}
\]
where the pre‑soft‑max score  
\(
z_{ij}= \tfrac{1}{n}\mathbf 1_n^{\!\top}\!
        \underbrace{\hat A H_{i-1}^{t}W_i^{t}W_{i}^{pt}}_{\displaystyle T_i^{p}}
        (W_j^{ps})^{\!\top} (W_j^{s})^{\!\top} (H_{j-1}^{s})^{\!\top}\hat A^{\!\top}
        \mathbf 1_n/n
\)  
is a scalar obtained by taking the trace of the product of two
length‑\(n\) vectors (so dimensions always match).  \(M_{ij}\) is the
corresponding dissimilarity entry of \(\mathbf M\).

By the product rule
\[
\frac{\partial L_{ij}}{\partial W_j^{s}}
  = M_{ij}\frac{\partial \alpha_{ij}}{\partial W_j^{s}}
   +\alpha_{ij}\frac{\partial M_{ij}}{\partial W_j^{s}}.
\tag{A2}
\]

Because \(\alpha_{ij}\) is a soft‑max,  
\(
\frac{\partial \alpha_{ij}}{\partial z_{i'j}}
   =\alpha_{ij}(\delta_{ii'}-\alpha_{i'j}).
\)  
Applying the chain rule,
\begin{align}
\frac{\partial \alpha_{ij}}{\partial W_j^{s}}
   =\!\sum_{i'=1}^{T_\ell}
     \frac{\partial \alpha_{ij}}{\partial z_{i'j}}
     \frac{\partial z_{i'j}}{\partial W_j^{s}}
   \;=\!\sum_{i'=1}^{T_\ell}
     \alpha_{ij}\bigl(\delta_{ii'}-\alpha_{i'j}\bigr) \\
     \bigl[\, (H_{j-1}^{s})^{\!\top}\hat A^{\!\top}\tfrac{\mathbf 1_n}{n}\bigr]\,
     \bigl[\tfrac{\mathbf 1_n^{\!\top}}{n}\,T_{i'}^{p}(W_j^{ps})^{\!\top}\bigr].
\tag{A3}
\end{align}
Each factor \(T_{i'}^{p}=\hat A H_{i'-1}^{t}W_{i'}^{t}W_{i'}^{pt}\)
contains the teacher weight matrix \(W_{i'}^{t}\).  Therefore
\(
\partial\alpha_{ij}/\partial W_j^{s}
\) depends on \emph{every} \(W_{i'}^{t}\).
Both \(D_{\text{emb},ij}\) and \(D_{\text{str},ij}\) are functions of
\(T_i\) and \(S_j\); their gradients w.r.t.\ \(W_j^{s}\) pass through
\(T_i\) exactly once, so
\(
\partial M_{ij}/\partial W_j^{s}
\) also carries \(W_i^{t}\).

The STRIDE loss is the average over all \((i,j)\):
\(
L_{\text{STRIDE}}=\tfrac{1}{S_\ell}\sum_{i,j} L_{ij}.
\)
Summing Eq.(A2) over \(i\) preserves the dependence on every
teacher weight appearing in (A3).  Hence
\(
\partial L_{\text{STRIDE}}/\partial W_j^{s}
\)
is a function of the whole set \(\{W_i^{t}\}_{i=1}^{T_\ell}\). Since the argument holds for any student layer \(j\), the gradient
for \emph{every} student layer jointly involves \emph{all} teacher
layers, completing the proof.

\vspace{-2mm}

\section{Summary of Notations}
\vspace{-4ex}
In the ``Proposed'' section, we mathematically describe how \system generates the attention matrix, $A$, the structural dissimilarity matrix, $Q$, and the embedding dissimilarity matrix, $D$, which is then used to calculate $L_{\system}$. In Table \ref{tab:notation}, we provide a summary of all the mathematical notation used to describe \system.

% =========================  NOTATION TABLE  =============================
\begin{table}[H]
%\vspace*{-4ex}
\centering\small
\resizebox{0.45\textwidth}{!}{
\begin{tabular}{l l}
\toprule
\textbf{Symbol} & \textbf{Meaning / Shape} \\
\midrule
\multicolumn{2}{c}{\emph{Graph primitives}} \\
$n$ & Number of nodes in the input graph \\
$\mathbf{A}\in\{0,1\}^{n\times n}$ & Binary adjacency matrix \\
$\mathbf{D}=\mathrm{diag}(d_1,\dots,d_n)$ & Degree matrix ($d_i=\sum_j\!A_{ij}$) \\
$\hat{\mathbf A}= \mathbf D^{-1/2}\mathbf A\mathbf D^{-1/2}$ & Symmetric normalised adjacency (Eq. 2) \\
$\mathbf 1_{n}\in\mathbb R^{n}$ & Vector of ones (all entries = 1) \\
\midrule
\multicolumn{2}{c}{\emph{Index sets and dimensions}} \\
$T_\ell,\;S_\ell$ & Number of layers in teacher / student networks \\
$i\in\{1,\dots,T_\ell\}$ & Teacher‑layer index \\
$j\in\{1,\dots,S_\ell\}$ & Student‑layer index \\
$d_t,\;d_s$ & Hidden dimension of a teacher / student layer \\
$d_t^{(i)},\;d_s^{(j)}$ & Output dimension of teacher layer $i$ / student layer $j$ \\
$d_l$ & Hidden dimension at generic layer $l$ \\
$d_a$ & STRIDE latent dimension for projections / attention \\
\midrule
\multicolumn{2}{c}{\emph{Layer outputs and projections}} \\
$\mathbf H^{(l)}\in\mathbb R^{n\times d_l}$ & Node‑feature matrix at layer $l$ \\
$\mathbf T_i=\mathbf H^{t}_{i}\in\mathbb R^{n\times d_t}$ & Pre‑activation output of teacher layer $i$ \\
$\mathbf S_j=\mathbf H^{s}_{j}\in\mathbb R^{n\times d_s}$ & Pre‑activation output of student layer $j$ \\
$\mathbf T_i^{p} = \hat{\mathbf A} \mathbf H^{t}_{i-1} W_i^t W_i^{pt}$ & Projected teacher representation (Eq. 3) \\
$\mathbf S_j^{p} = \hat{\mathbf A} \mathbf H^{s}_{j-1} W_j^s W_j^{ps}$ & Projected student representation (Eq. 3) \\
$\mathbf T_i^{p},\;\mathbf S_j^{p}\in\mathbb R^{d_a}$ & Mean‑pooled projected representations \\
\midrule
\multicolumn{2}{c}{\emph{Trainable weight matrices}} \\
$W_i^{t}\in\mathbb R^{d_{t}^{(i-1)}\times d_{t}^{(i)}}$ & GNN weight matrix of teacher layer $i$ \\
$W_j^{s}\in\mathbb R^{d_{s}^{(j-1)}\times d_{s}^{(j)}}$ & GNN weight matrix of student layer $j$ \\
$W_{i}^{pt}\in\mathbb R^{d_t\times d_a}$ & Teacher projection for attention (layer $i$) \\
$W_{j}^{ps}\in\mathbb R^{d_s\times d_a}$ & Student projection for attention (layer $j$) \\
$P_t\in\mathbb R^{d_t\times d_a},\;P_s\in\mathbb R^{d_s\times d_a}$ & Embedding dissimilarity projections \\
$E_t\in\mathbb R^{d_t\times d_a},\;E_s\in\mathbb R^{d_s\times d_a}$ & Structural dissimilarity projections \\
$\mathbf P\in\mathbb R^{d_a\times d_a}$ & Shared low‑rank sub‑space projection (§3) \\
\midrule
\multicolumn{2}{c}{\emph{Attention and dissimilarity tensors}} \\
$\alpha_{ij}$ & Attention score for teacher layer $i$ $\leftrightarrow$ student layer $j$ \\
$\boldsymbol{\alpha}\in\mathbb R^{T_\ell\times S_\ell}$ & Full attention matrix (Eq. 3) \\
$z_{ij}$ & Scalar pre‑soft‑max compatibility score for pair $(i,j)$ \\
$D^{\text{emb}}_{ij}$ & Pairwise \emph{embedding} dissimilarity (Eq. 4) \\
$D^{\text{str}}_{ij}$ & Pairwise \emph{structural} dissimilarity (Eq. 5) \\
$\mathbf D_{\text{emb}},\;\mathbf D_{\text{str}}\in\mathbb R^{T_\ell\times S_\ell}$ & Two dissimilarity matrices \\
$M_{ij}=D^{\text{emb}}_{ij}+D^{\text{str}}_{ij}$ & Total dissimilarity for pair $(i,j)$ \\
$\mathbf M=\mathbf D_{\text{emb}}+\mathbf D_{\text{str}}$ & Total dissimilarity matrix \\
\midrule
\multicolumn{2}{c}{\emph{Losses, operators and hyper‑parameters}} \\
$L_{ij}= \alpha_{ij} M_{ij}$ & STRIDE loss contribution of a single layer pair \\
$L_{\text{STRIDE}}$ & Global STRIDE distillation loss (Eq. 6) \\
$\mathcal H(\cdot,\cdot)$ & Cross‑entropy loss (logit supervision) \\
$\sigma(\cdot)$ & Element‑wise activation function \\
$\phi(\cdot,\cdot)$ & G‑CRD structural contrastive loss (Eq. 5) \\
$\odot$ & Hadamard (element‑wise) product \\
$\beta$ & Trade‑off coefficient in the total loss (Eq. 7) \\
\bottomrule
\end{tabular}
}
\caption{\small Comprehensive notation used throughout the STRIDE paper. Bold uppercase symbols denote matrices, bold lowercase symbols denote vectors, and plain symbols denote scalars unless stated otherwise. Dimensions are provided where applicable.}
\label{tab:notation}
\end{table}

\end{document}